\documentclass[final,5p,times,twocolumn,authoryear]{elsarticle}

\usepackage{natbib}
\usepackage{amsmath,amsfonts,amssymb} 
\usepackage{amsthm}
\usepackage{algorithmic}
\usepackage{algorithm}
\usepackage{tabularx} 
\usepackage{array}
\usepackage{xcolor}   
\usepackage{textcomp}
\usepackage{stfloats}
\usepackage{verbatim}
\usepackage{graphicx}
\usepackage{multirow}
\usepackage{booktabs, makecell}    
\usepackage{nicefrac}       
\usepackage{microtype}      
\usepackage{subfig}
\usepackage{bm}
\usepackage[capitalize,noabbrev]{cleveref}

\begin{document}

\begin{frontmatter}




\title{Jingfang: An LLM-Based Multi-Agent System for Precise Medical Consultation and Syndrome Differentiation in Traditional Chinese Medicine}



\author[1,3]{Yehan Yang\fnref{lab1}}
\author[2]{Tianhao Ma\fnref{lab1}}
\author[1]{Ruotai Li\corref{cor1},\fnref{lab1}}
\ead{rtli@bupt.edu.cn}
\author[1,4]{Xinhan Zheng}
\author[1]{Guodong Shan}
\fntext[lab1]{These authors contributed equally to this work.}
\cortext[cor1]{Corresponding author}

\affiliation[1]{organization={Beijing University of Posts and Telecommunications},
            addressline={No.10 Xitucheng Road,Haidian District }, 
            city={Beijing},
            postcode={100876}, 
            state={},
            country={China}}
 \affiliation[2]{organization={Southeast University},
            addressline={No. 2, Southeast University Road, Jiangning District}, 
            city={Nanjing},
            postcode={211189}, 
            state={Jiangsu},
            country={China}}           
\affiliation[3]{organization={University of Chinese Academy of Sciences},
            addressline={No.1 East Yanqi Lake Road,Huairou District}, 
            city={Beijing},
            postcode={101408}, 
            state={},
            country={China}}
\affiliation[4]{organization={University of Science and Technology of China},
            addressline={No.100 Fuxing Road}, 
            city={Hefei},
            postcode={230026}, 
            state={Anhui},
            country={China}}


\begin{abstract}

The practice of Traditional Chinese Medicine (TCM) requires profound expertise and extensive clinical experience. While Large Language Models (LLMs) offer significant potential in this domain, current TCM-oriented LLMs suffer two critical limitations: (1) a rigid consultation framework that fails to conduct comprehensive and patient-tailored interactions, often resulting in diagnostic inaccuracies; and (2) treatment recommendations generated without rigorous syndrome differentiation, which deviates from the core diagnostic and therapeutic principles of TCM. To address these issues, we develop \textbf{JingFang (JF)}, an advanced LLM-based multi-agent system for TCM that facilitates the implementation of AI-assisted TCM diagnosis and treatment. JF integrates various TCM Specialist Agents in accordance with authentic diagnostic and therapeutic scenarios of TCM, enabling personalized medical consultations, accurate syndrome differentiation and treatment recommendations. A \textbf{Multi-Agent Collaborative Consultation Mechanism (MACCM)} for TCM is constructed, where multiple Agents collaborate to emulate real-world TCM diagnostic workflows, enhancing the diagnostic ability of base LLMs to provide accurate and patient-tailored medical consultation. Moreover, we introduce a dedicated \textbf{Syndrome Differentiation Agent} fine-tuned on a preprocessed dataset, along with a designed \textbf{Dual-Stage Recovery Scheme (DSRS)} within the Treatment Agent, which together substantially improve the model's accuracy of syndrome differentiation and treatment. Comprehensive evaluations and experiments demonstrate JF's superior performance in medical consultation, and also show improvements of at least 124\% and 21.1\% in the precision of syndrome differentiation compared to existing TCM models and State of the Art (SOTA) LLMs, respectively. 

\end{abstract}




\end{frontmatter}



\section{Introduction}
\label{Introduction}
In recent years, the emergence of large language models (LLMs) such as GPT-5.2 \cite{gpt5.2}, Llama3 \cite{2}, Deepseek-MOE~\cite{3}, Deepseek v3.2~\cite{25}, Gemini-3~\cite{31} and Qwen3 \cite{4} has significantly advanced the field of natural language processing. These models not only possess exceptional language comprehension and generation capacity~\cite{21}, but also demonstrate strong cross-domain adaptability and applicability~\cite{5,6,7} by integrating techniques such as pre-training~\cite{16}, fine-tuning~\cite{17}, reinforcement learning~\cite{18}, human feedback alignment~\cite{19}, and knowledge graph retrieval~\cite{20}. Among the numerous cross-domain applications of LLM, TCM LLM has become one of the key focuses due to the importance of TCM in health care and disease treatment~\cite{23,24}. Due to the complexity of medical consultations and the need for profound professional knowledge in diagnosis and treatment of TCM, the effective application of LLMs and related technologies in the clinical practice of TCM poses a significant challenge.

\subsection{Related Work} 
Recently, several studies that incorporate LLM and related technologies have been conducted to promote the practice of AI-assisted TCM and have made some meaningful attempts. For example: TCMChat~\cite{9} is a generative LLM tailored for TCM, built on Baichuan2-7B-Chat and trained on diverse TCM data across six task categories. It excels in performing TCM-related question-answering and diagnostic tasks. ZhongJing~\cite{8} is characterized by its integration of reinforcement learning with human feedback, combined with CMtMedQA, which improves proactive reasoning and knowledge alignment. HuaTuo~\cite{27} features an innovative one-stage domain adaptation protocol that unifies heterogeneous data into pairs of instruction-output format and introduces a data priority sampling strategy. ShenNong~\cite{28} adopts a framework that facilitates the performance of LLM on TCM tasks using limited data through a two-stage approach of supervised fine-tuning with medical cases followed by reinforcement learning from AI feedback. SunSimiao~\cite{29} is a Qwen2-7B fine-tuned model using high-quality medical data, achieving good performance equivalent to 30B scale models in CMB-Exam.

Although these interesting and meaningful attempts have been made, the critical limitations of the effective application of existing TCM LLMs in real clinical practice remain. The fundamental paradigm of TCM diagnosis and treatment emphasizes the differentiation of the syndrome in diagnosis and then provides the syndrome-specific treatment recommendation. Accurate and comprehensive medical consultation lays the foundation for a precise diagnosis and treatment, which requires the practitioner of TCM to conduct multiple rounds of personalized and targeted inquiries and interactions with patients. In the process of medical consultation, professional TCM doctors carefully collect useful information from patients, analyze their symptoms and signs, attribute them to the corresponding TCM syndromes, and provide effective TCM treatment recommendations accordingly. The entire process is a typical procedure for the differentiation and treatment of the TCM syndrome.

Clinical medical consultations and diagnostic processes in real world settings are highly variable and patient-specific. This inherent complexity makes it difficult for existing TCM LLMs to accurately simulate and align with authentic TCM diagnostic consultation scenarios solely through the basic language processing capabilities of LLMs or predefined consultation templates. The scarcity of large-scale, well-annotated multi-round consultation data hinders the training of TCM LLMs to achieve accurate diagnostic consultation capabilities. Consequently, current TCM LLMs lack the expert-level precision and targeted medical consultation capabilities required for alignment with real-world TCM clinical applications. This deficiency leads to an incomprehensive and imprecise model-based diagnosis, which directly affects the precision of subsequent syndrome differentiation and corresponding treatment. Furthermore, existing TCM LLMs often fail to perform precise syndrome differentiation, and some even provide treatment recommendations without an explicit syndrome basis. This limitation causes TCM LLMs to lack interpretability and reliability in diagnosis and treatment, directly affecting their precision of treatment and seriously hindering their clinical practice.

\subsection{Main Contribution} 
To address these, (1) we develop an \textbf{advanced LLM-based multi-agent system}, named JingFang (JF), that addresses the limitations of existing models, especially in \textbf{accurate medical consultation, diagnosis, and syndrome differentiation}, directly facilitating the effective applications of LLM and related technologies in TCM. The JF system is meticulously developed to incorporate a range of TCM Specialist Agents, with the objective of simulating and aligning with authentic diagnostic and therapeutic scenarios of TCM to the greatest extent. (2) A \textbf{Multi-Agent Collaborative Consultation Mechanism (MACCM)} for TCM is designed to emulate the real-world TCM consultation workflow through the effective collaboration of multiple Agents. This enables comprehensive and patient-tailored medical consultations and improves the diagnostic accuracy of base LLMs. (3) A \textbf{Syndrome Agent} has been systematically trained based on a preprocessed structured TCM dataset, and a \textbf{Dual-Stage Retrieval Scheme (DSRS)} is proposed for the Treatment Agent, which substantially improves the interpretability and reliability of the model in precise syndrome differentiation and the corresponding treatment.

\subsection{Organization} 
In Section 2,  we introduce the approaches for the construction of functional modules in the framework, the design and development of various TCM Specialist Agents, and the training method of the Jingfang model. In Section 3, the superior performance of the JF system in accurate medical consultation and syndrome differentiation is demonstrated through comprehensive experiments and evaluations. Ablation experiments and case studies have also been conducted to show the effectiveness and reliability of our methods. Some conclusions and discussions are given in Section 4. 


\section{Method}
\label{Method} 
\subsection{Framework of JingFang}
\begin{figure}[htbp]
\begin{center}
\centerline{\includegraphics[width=0.95\columnwidth]{Figure_pdf/JingFang.pdf}}
\caption{Overall workflow framework of JingFang, including three main modules: TCM Consultation, TCM Syndrome Differentiation, and TCM Treatment, which enables the model to simulate the real-world diagnosis and treatment scenarios of TCM.}
\label{jingfang-Framwork}
\end{center}
\end{figure}
As illustrated in Fig.~\ref{jingfang-Framwork}, JF system follows the authentic clinical workflow of TCM and includes three core modules: \textbf{TCM Consultation}, \textbf{TCM Syndrome Differentiation}, and \textbf{TCM Treatment Recommendation}. In the \textbf{Consultation Module}, the MACCM integrates multiple TCM Specialist Agents and the TCM General Agent to realize expert-level multi-round consultations. This mechanism enables the model to adaptively conduct patient-based medical consultation and collect comprehensive information during the consultation process. In the \textbf{Syndrome Differentiation Module}, the Syndrome Agent is fine-tuned on a high-quality structured TCM syndrome dataset and performs syndrome differentiation based on the classical ``Ten Questions Song'' framework of TCM, ensuring both diagnostic accuracy and reasoning interpretability. In the \textbf{Treatment Recommendation Module}, the TCM Treatment Agent is developed by adopting the DSRS that integrates sparse lexical and dense semantic embedding–based retrieval, to systematically identify and rank prescriptions from a structured TCM formula dataset, ensuring precise and syndrome-specific therapeutic recommendations. Together, these three modules form an intelligent closed-loop multiple Agents system that aligns with the real clinical scenario of TCM, achieving the full workflow of \emph{consultation–syndrome differentiation–treatment} in a cohesive and interpretable manner.
\subsection{Multi-Agent Collaborative Consultation Mechanism}
After an in-depth study of the real-world TCM consultation and diagnosis process, we found that experienced TCM experts typically pre-set one or two follow-up questions based on the patient's known conditions (chief complaint) and make adaptive and targeted inquiries to collect more information during the consultation process. Inspired by this, the Multi-Agent Collaborative Consultation Mechanism (MACCM) is developed to better fit the authentic TCM consultation process, and an TCM Agents group $\mathcal{E}$ is constructed accordingly, where the prompts: $prompt_{[\ ]}$ are particularly designed to facilitate different TCM Specialist Agents in implementing a specific function. This group consists of \textbf{TCM Specialist Agents} from multiple clinical domains such as \textbf{internal medicine, surgery, gynecology, and pediatrics}, and can be formally described as
\begin{equation}
    \mathcal{E} = \{\, \{E_1,K^{spec}_1\}, \{E_2,K^{spec}_2\}, \cdots, \{E_i,K^{spec}_i\},\cdots \,\},
\end{equation}
where each Specialist Agent $E_i$ is associated with its corresponding domain-specific knowledge, denoted by $K^{spec}_{i}$. 

\subsubsection{Generation of TCM Agents Team} 
At the beginning of the consultation, based on the patient's chief complaint $M_{comp}$, the Manager Agent selects the most appropriate TCM Specialist Agents from the Agents group $\mathcal{E}$ to generate a patient-targeted Agents team for the current consultation. This patient-based agent-driven selection and construction process can be formalized as
\begin{equation}
    SD_i = {Agent_{select}}\bigl(M_{comp},\ \mathcal{E},\ \mathrm{prompt}_{select}\bigr),
\end{equation}
where $Agent_{select}$ represents the agent-driven mechanism that integrates the patient's chief complaint, the Agents group $\mathcal{E}$, and the specialized prompts to generate the patient-based TCM Specialist Agent $SD_i$.

To further ensure the comprehensiveness of the consultation process, a \textbf{TCM General Agent} is also included, which is defined as $GD = \{ E^{gen},\ K^{gen} \}$, denoting a general-purpose TCM expert $E^{gen}$ equipped with fundamental and general TCM knowledge $K^{gen}$. Therefore, the final TCM patient-based Agent team can be constructed as
\begin{equation}
    Team(M) = \{ SD_i,\ GD \},
\end{equation}
which enhances the targeted and comprehensive medical consultation. It should be noted that the TCM Agent team for each patient is established only once during the consultation process.

\subsubsection{Construction of Consultation} 
After the t-th interaction with the patient, the TCM Record Agent summarizes the \textbf{patient’s current condition $ R_t$} from the previous consultation result, which is defined as: $C_t = \{(Q_k, A_k) \mid k \in [0, t]\}$, where $Q_k$ and $A_k$ are the inquiry and the corresponding answer at the k-th round of consultation. Thus, $R_t$ can be defined as:
\begin{equation}
   R_t = Agent_{record}(C_t,\ prompt_{record}).
\end{equation}
To formally initiate the (t+1)-th consultation, each Agent in the TCM Agent team first prioritizes one or two subsequent consultation questions based on the current result $C_t$, providing with explanations and theoretical guidance. Specifically, the TCM Specialized Agent $SD_i$ leverages its domain-specific professional knowledge, together with the specialized prompt $prompt_{spec}$, to generate patient-targeted consultation questions $Q^{spec}_{i,t+1}$ for the (t+1)-th round of consultation:
\begin{equation}
    Q^{spec}_{i,t+1} = Agent_{spec}(C_t,\ SD_i,\ prompt_{spec}).
\end{equation}
Meanwhile, the TCM General Agent $GD$ generates more general but comprehensive questions $Q^{gen}_{t+1}$ from the perspective of fundamental knowledge of TCM :
\begin{equation}
    Q^{gen}_{t+1} = Agent_{gen}(C_t,\ GD,\ prompt_{gen}).
\end{equation}
The initial set of consultation questions for the (t+1)-th round is constructed by combining the specialized questions $Q^{spec}_{i,t+1}$ and the general questions $Q^{gen}_{t+1}$:
\begin{equation}
    Q^{init}_{t+1} = (Q^{spec}_{i,t+1},\ Q^{gen}_{t+1}).
\end{equation}
Subsequently, the Agents team iteratively refines and optimizes $Q^{init}_{t+1}$ to progressively improve the precision of the content, ultimately determining the final question to be posed to the patient in this round.

\subsubsection{Evaluation of Consultation} 
To facilitate the generation of the final formal consultation questions for this round, an accurate and comprehensive evaluation is required. Here, we utilize the Ten Questions Song~\cite{26} (TQS), which is a widely used classical framework in authentic TCM consultation, to establish a Consultation Questions Evaluation Algorithm (CQEA) for the Evaluation Agent.  
\begin{figure}
  \setlength{\intextsep}{0pt} 
  \centering
  \begin{minipage}{\linewidth}
    \begin{algorithm}[H]
    \caption{Consultation Questions Evaluation Algorithm}
    \label{alg:ConsultationCoT}
    \begin{algorithmic}[1]
    \STATE \textbf{Initialization:} $scored\_consultation \leftarrow \emptyset$
    \FOR{$question$ in ${Q^{eval}_{t+1,j}}$}
        \STATE $com\_score,\ per\_score \leftarrow 0,\ 0$ 
        \FOR{item in $ten\_questions\_song$}
            \STATE $score \leftarrow sim(emb(question),\ emb(item))$ 
            
            \IF{$score > com\_score$}
                \STATE \quad $com\_score \leftarrow score$
            \ENDIF
        \ENDFOR
        
        \FOR{$item$ in ${core\_questions, medical\_record}$}
            \STATE $score \leftarrow sim(emb(question),\  emb(item))$ 
            
            \IF{$score > per\_score$}
                \STATE \quad $per\_score \leftarrow score$ 
            \ENDIF
        \ENDFOR
        \STATE$total\_score \leftarrow com\_score + per\_score$ 
        \STATE \textbf{Store:} Add the $question$ with its $total\_score$ to the $scored\_consultation$ 
    \ENDFOR
    \STATE \textbf{Return:} $scored\_consultation$ 
    \end{algorithmic}
    \end{algorithm}
  \end{minipage}
\end{figure}

According to the CQEA, the Evaluation Agent utilizes an embedding model to calculate the similarity between the generated questions and those of the TQS, thus obtaining a comprehensiveness score. The pertinence score of the generated questions is achieved by incorporating the key issues of each Specialist Agent (each Agent focuses on characteristic diseases and diagnostic points related to their specialty) and the patient's known condition to calculate their similarity. The final evaluation score is obtained by aggregating these two scores. Therefore, this evaluation-and-optimization iterative process that facilitates the generation of formal consultation questions for this round can be described as follows. At any j-th sub-iteration turn, the Evaluation Agent generates the consultation questions $Q^{eval}_{t+1,j}$ that include a summary and evaluation through the knowledge-based CQEA, based on the current patient condition $R_t$ and the questions given by the Optimization Agent at (j-1)-th turn:
\begin{equation}
    Q^{eval}_{t+1,j} = Agent_{eval}(Q^{opt}_{t+1,j-1},\ R_t,\ prompt_{eval}).
\end{equation}

Especially, when the initial $j=1$, set $Q^{opt}_{t+1,0}=Q^{init}_{t+1}$.

\subsubsection{Optimization of Consultation} 
After obtaining the question evaluation in the j-th sub-iteration turn, each Agent in the TCM Agents team independently assesses the evaluated consultation questions $Q^{eval}_{t+1,j}$ based on their domain expertise and the current state $R_t$, providing their individual modification suggestions $Mod^{(i)}_{t+1,j}$. An Optimization Agent is designed to integrate the $R_t$, $Q^{eval}_{t+1,j}$ and the all modification suggestions $Mod^{(i)}_{t+1,j}$, generating the optimized consultation questions for this sub-iteration turn: 

\begin{equation}
Q^{opt}_{t+1,j}
=
Agent_{opt}
\left(
Q^{eval}_{t+1,j},
\sum_{i=1}^{|\mathrm{Team}|} Mod^{(i)}_{t+1,j},
R_t,
prompt_{opt}
\right).
\end{equation}

These optimized consultation questions are then proposed to the Evaluation Agent for the next turn evaluation. This evaluation-and-optimization iterative process continues until a consensus is reached among all Agents or the maximum number of feedback turns has been reached. The final consultation question, $Q^{final}_{t+1}$, is then generated and could be used to initiate the formal consultation with the patient for the current round.

\subsubsection{Multi-Round Consultation Driven by Multi-Agent Collaboration} 
Based on the final consultation question $Q^{final}_{t+1}$ and the current consultation result $C_t$ in the t-$th$ round, a Consultation Agent is developed following real-world consultation scenarios to interact with the patient and obtain the consultation result:
\begin{equation}
    C_{t+1}=Agent_{consult}(Q^{final}_{t+1}, \ C_t,\ prompt_{consult}).
\end{equation}
Then, the Record Agent is called to systematically update and integrate the patient's condition into the medical record $R_{t+1}$. As the process continues, a series of targeted questions is generated to further explore the patient's condition and progressively gather more comprehensive information. Once the necessary information has been collected or the maximum number of consultation rounds has been reached, the Record Agent will present the final medical case $R_f$ to the Syndrome Differentiation and Treatment Recommendation Modules. Fig.\ref{MACCM} shows an example of how this MACCM works in a medical consultation scenario. 
\begin{figure}
\setlength{\intextsep}{0pt} 
\centering
\includegraphics[width=1\columnwidth]{Figure_pdf/An_example_of_multi_agent_simulation_of_TCM_consultation_CoT.pdf}
\caption{An example of the workflow of the proposed MACCM that enhances the accurate and targeted medical consultation.}
\label{MACCM}
\end{figure}

\subsection{TCM Syndrome Agent} 
In TCM, precise syndrome differentiation is essential for effective treatment, which requires high-quality datasets for LLM training. However, we found that real-world TCM records contain substantial noise that is resistant to conventional regularization. This noise includes irrelevant administrative statements, such as ``patient visited for integrated treatment'', and references to Western medical examinations, such as ``patient refused cranial CT'', which distract models from essential disease information. Content based on instrumental diagnostics, e.g. ``blood amylase tests normal, abdominal CT shows colon wall thickening'', must also be filtered out, in order to maintain a focus on TCM-relevant symptom patterns.

In real-world TCM scenarios, TCM experts generally utilize TQS to fully understand the patient's symptoms, medical history, and physical signs through a series of fundamental questions, obtaining key information for a precise diagnosis. Inspired by this, we propose a universal data preprocessing method using LLM to automatically extract key information related to the TQS from raw data and strictly retain the core information closely related to the patient’s condition. This ensures alignment between the output of the case in the JF and the format of the training medical case data. The systematic preprocessing of data has laid a solid foundation for subsequent accurate diagnosis and precise syndrome differentiation, significantly enhancing the model's ability of syndrome differentiation-based treatment.

Based on the cleaned and structured dataset of 43,000 high-quality cases, we fine-tuned two models to allow JingFang’s precise syndrome differentiation capability: 
(1) a \textbf{LoRA-tuned Qwen2.5-7B-Instruct} model (rank = 64, $\alpha$ = 16, learning rate = $5 \times 10^{-4}$, batch size = 16), and 
(2) a \textbf{RoBERTa-based} classifier with an additional multi-classification head trained with mixed precision fp16 for efficiency. All training was performed on four NVIDIA A6000 GPUs (48GB each), ensuring stable convergence of the loss function and high computational efficiency.

Once the final patient record, $R_f$, has been produced after the multi-round consultation, the TCM Syndrome Agent automatically analyzes it, combining the diagnostic reasoning prompt, $prompt_s$, and inferring the most probable type of syndrome:
\begin{equation}
    TCM_{sy} = Agent_{sy}(R_f, prompt_{sy}).
\end{equation}

\subsection{TCM Treatment Agent}
TCM treatment is formulated based on the patient’s syndrome, while also integrating various aspects of patient information, such as the chief complaints and medical history, to dynamically adjust treatment strategies and provide precise, personalized care. To enable the TCM Treatment Agent to do this, we propose a \textbf{ Dual-Stage Retrieval Scheme (DSRS)} that automatically retrieves and recommends the most suitable prescriptions from a structured TCM prescription database.

The DSRS divides the retrieval process into two stages. In the \textbf{first stage}, the database is filtered according to syndrome-related features (e.g., etiology, affected organ, and type of syndrome) to quickly obtain candidate prescriptions. In the \textbf{second stage}, sparse and dense semantic embeddings are used together with a  Reciprocal Rank Fusion (RRF) algorithm to re-rank the results, balancing lexical precision with semantic relevance. The system then outputs the three most similar prescriptions (TOP-3) as the final recommendations.

The generation process is outlined as follows:
\begin{equation}
    TCM_{tr} = Agent_{tr}(R_f, DB_{TCM}, prompt_{tr})
\end{equation}
where $DB_{TCM}$ represents a structured TCM prescription database that covers internal medicine, gynecology, pediatrics, and surgery. 
Each entry includes essential information such as disease categories, syndrome types, clinical manifestations, representative formulas, commonly used herbs, and associated therapeutic methods.

\begin{algorithm}[htbp]
\caption{Dual-Stage Retrieval Scheme (DSRS) for TCM Treatment Recommendation}
\label{alg:dsrs}
\begin{algorithmic}[1]
\STATE \textbf{Input:} Final medical record $R_f$, TCM prescription database $DB_{TCM}$, prompt $p$
\STATE \textbf{Stage 1: Syndrome-Based Filtering}
\STATE Extract syndrome attributes (type, etiology, affected organ) from $R_f$
\STATE Filter $DB_{TCM}$ to obtain candidate set $C$
\STATE \textbf{Stage 2: Hybrid Ranking}
\STATE Encode $R_f$ and $C$ using sparse lexical embeddings $E_s$ and dense semantic embeddings $E_d$
\STATE Conduct searches to obtain ranked lists $R_s$ (sparse) and $R_d$ (dense)
\STATE Fuse rankings using Reciprocal Rank Fusion (RRF) to produce $R_{hyb}$
\STATE \textbf{Output:} Return the Top-3 prescriptions from $R_{hyb}$ as final recommendations
\end{algorithmic}
\end{algorithm}

\section{Experiments}
In this section, we present a series of systematic experiments designed to assess the performance of JingFang (JF) in two core aspects: the precision of TCM syndrome differentiation and the comprehensiveness and effectiveness of medical consultation. 

\subsection{Datasets and Baseline Models}
The dataset consists of three components: the TCM-SD dataset for syndrome differentiation \cite{mucheng-etal-2022-tcm}, 48 contemporary TCM textbooks encompassing internal medicine, surgery, gynecology, and pediatrics, and 700 historical TCM texts. The data has been meticulously reviewed and revised by three PhD students from various disciplines in Beijing University of Chinese Medicine. Thus, a high-quality TCM knowledge database has been constructed for the purpose of providing accurate treatment recommendations. This database contains 80,000 prescription entries, including 1,000 classic formulas. Each entry contains key attributes such as disease type, syndrome type, clinical manifestations, syndrome mechanism, treatment methods, representative formulas, and commonly used medicines. Furthermore, a total of 63,000 records pertaining to syndrome differentiation have been processed in their original form, and 43,000 samples of superior quality have been selected, accompanied by comprehensive case descriptions and corresponding diagnoses, for the purpose of model training on syndrome differentiation. 

To comprehensively evaluate the performance of JF, several representative open-source TCM models were selected as baselines. These models include ZhongJing-7B~\cite{8}, HuaTuo-32B~\cite{27}, ShenNong-7B~\cite{28}, and SunSimiao-7B~\cite{29}. Furthermore, four state-of-the-art (SOTA) large language models (LLMs) are incorporated: Claude-3.7~\cite{30}, DeepSeek v3~\cite{25}, Gemini-1.5-pro~\cite{31} and {GPT-4o}~\cite{hurst2024gpt} for comparison. A variety of evaluation metrics have been developed to ensure a comprehensive and reliable assessment. The utilization of task-specific evaluation criteria enables a multidimensional analysis of the performance of JF and other LLMs in different tasks.

\subsection{Evaluation of Syndrome Differentiation Precision}
The Qwen2.5-7B-Instruct and Roberta models are used as the foundation models for training the JF's syndrome differentiation ability, based on the curated TCM syndrome differentiation dataset. As Syndrome differentiation can also be considered a classification task, therefore, we trained it using these two different types of model. Training was conducted on four A6000 GPUs, each with 48 GB of VRAM. To evaluate the precision of syndrome differentiation, 8,699 TCM cases were randomly selected as the test dataset, covering 170 different syndrome types. Due to the imbalance in sample sizes across different categories, weighted evaluation metrics were used to assess the model’s performance to ensure a fair and accurate evaluation. The metrics include: \textbf{weighted precision}: $P_w$, \textbf{weighted recall}: $R_w$, and \textbf{weighted F1}: $F1_w$.

        
        
        
        

    
    
        
        

\begin{table}[htbp]
    \centering
    \renewcommand{\arraystretch}{1.3}
    \setlength{\tabcolsep}{12pt}
    \newcommand{\modelbox}[1]{\makebox[2.0cm][c]{#1}}
    \begin{tabular}{cccc}
        \toprule
        \modelbox{\textbf{Model}} & $P_w$ & $R_w$ & $F1_w$ \\
        \hline        

        \modelbox{\textbf{JF-RoBERTa}} 
        & \underline{0.8185} & \underline{0.8230} & \underline{0.8186} \\
        
        \modelbox{\textbf{JF-Qwen-2.5-7B}} 
        & \underline{0.8015} & \underline{0.8145} & \underline{0.8032} \\
        
        \modelbox{Claude-Opus-4.5} 
        & 0.6757 & 0.4286 & 0.5147 \\
        
        \modelbox{GPT-5.2} 
        & 0.6323 & 0.3029 & 0.3859 \\
        
        \modelbox{Gemini-3-Flash} 
        & 0.4597 & 0.3287 & 0.3499 \\
        
        \modelbox{DeepSeek-V3.2} 
        & 0.5091 & 0.3503 & 0.3814 \\
        
        \modelbox{HuaTuo} 
        & 0.3654 & 0.1735 & 0.1515 \\
    
        \modelbox{SunSimiao} 
        & 0.3113 & 0.1497 & 0.1415 \\
    
        \modelbox{ShenNong} 
        & 0.0713 & 0.0945 & 0.0401 \\
        
        \modelbox{ZhongJing} 
        & 0.0715 & 0.0598 & 0.0308 \\
        
        \Xhline{1pt}
    \end{tabular}
    \caption{Comparison of the performance of TCM syndrome differentiation among different models using weighted precision ($P_w$), weighted recall ($R_w$), and weighted F1-score ($F1_w$).}
    \label{tab:tcm_comparison_syndrome_differentiation}
\end{table}

As shown in Tab.~\ref{tab:tcm_comparison_syndrome_differentiation}, JF system outperforms both baseline TCM models and general state-of-the-art (SOTA) models in all evaluation metrics for the TCM syndrome differentiation task. Specifically, JF improves the precision of syndrome differentiation by at least 21.1\% compared to the general SOTA model (Claude-Opus-4.5), and by at least 124\% compared to the domain-specific TCM model (HuaTuo). Due to their significantly larger number of parameters compared to the baseline models of TCM, the general models Claude-Opus-4.5, GPT-5.2, Gemini-3-Flash and DeepSeek-V3.2 outperformed the TCM baseline models in certain metrics. However, they have not yet reached the precision of JF level. These results demonstrate the ability of our models to accurately differentiate TCM syndromes and the effectiveness of the TCM Multi-Agent Collaboration Mechanism that we have developed. It is important to note that the choice of foundation models does not significantly affect their capabilities in syndrome differentiation; rather, it is the training method and data that are key factors. Please refer to the ablation experiment in Sec.\ref{ablation experiment} for validation.

\subsection{Comprehensive Evaluation of Medical Consultation Ability}

\begin{figure*}[h]
  \centering
  \subfloat[Evaluation of Proactivity\label{fig:proactivity}]{
    \includegraphics[width=0.48\textwidth]{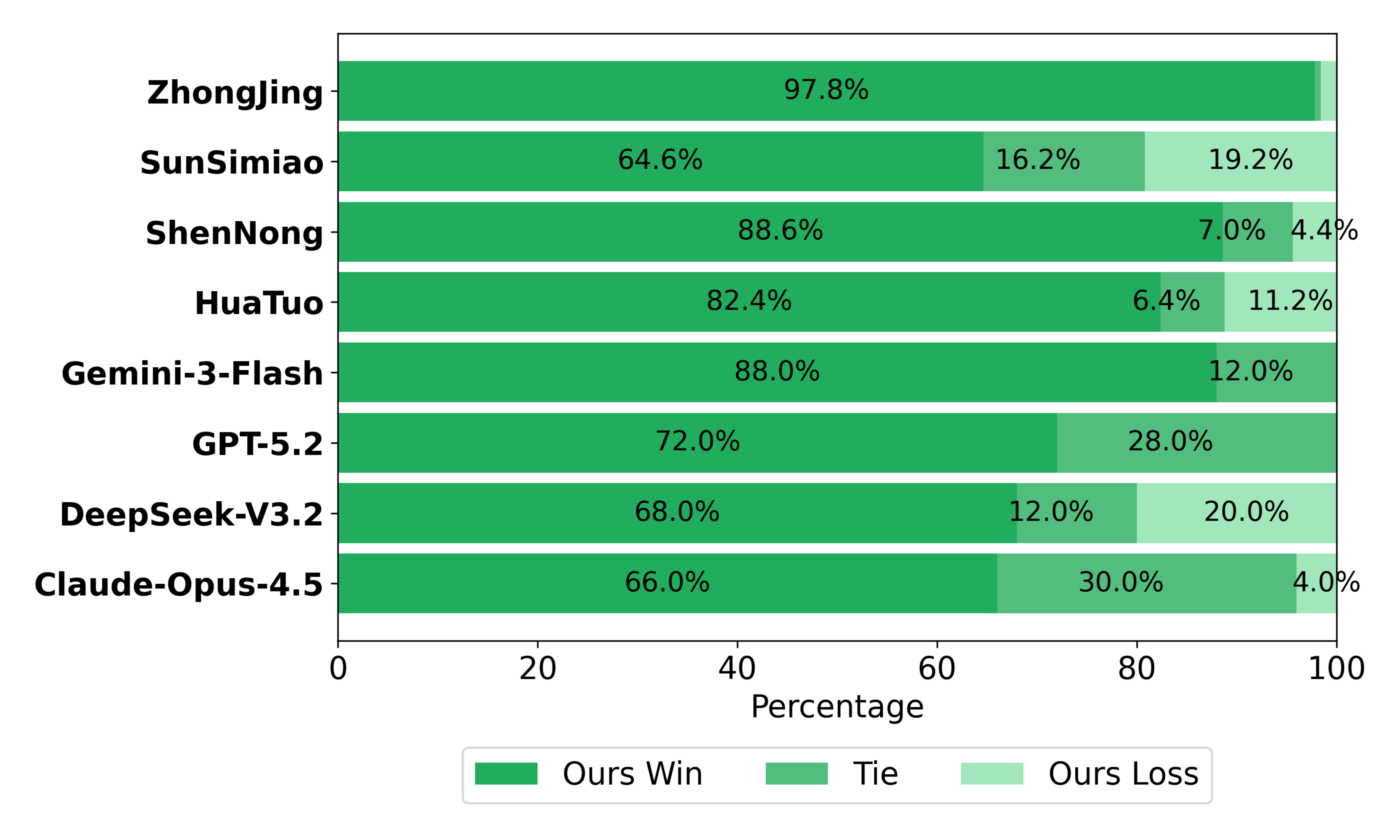}
  }\hfill
  \subfloat[Evaluation of Accuracy\label{fig:accuracy}]{
    \includegraphics[width=0.48\textwidth]{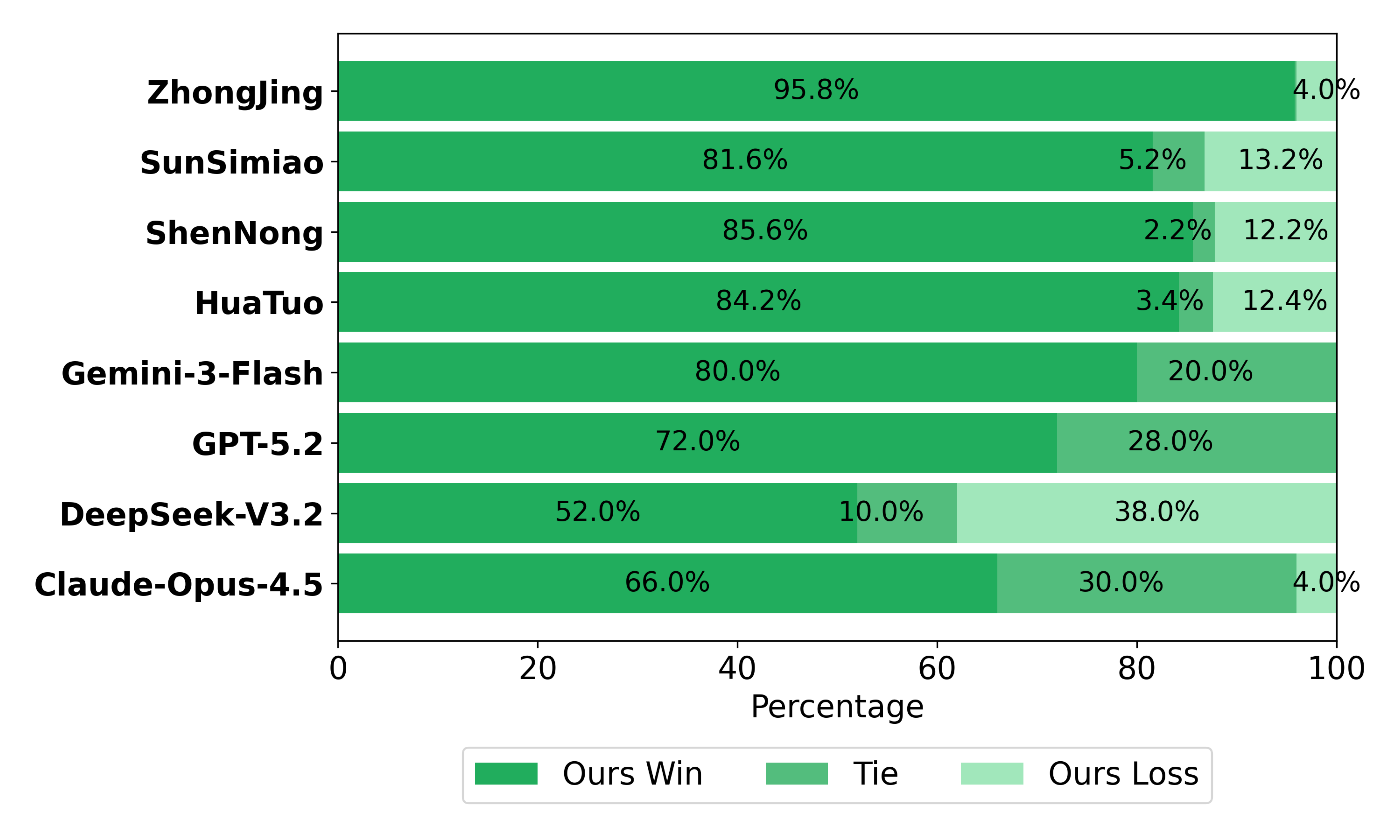}
  }\\ 
  \subfloat[Evaluation of Practicality\label{fig:practicality}]{
    \includegraphics[width=0.48\textwidth]{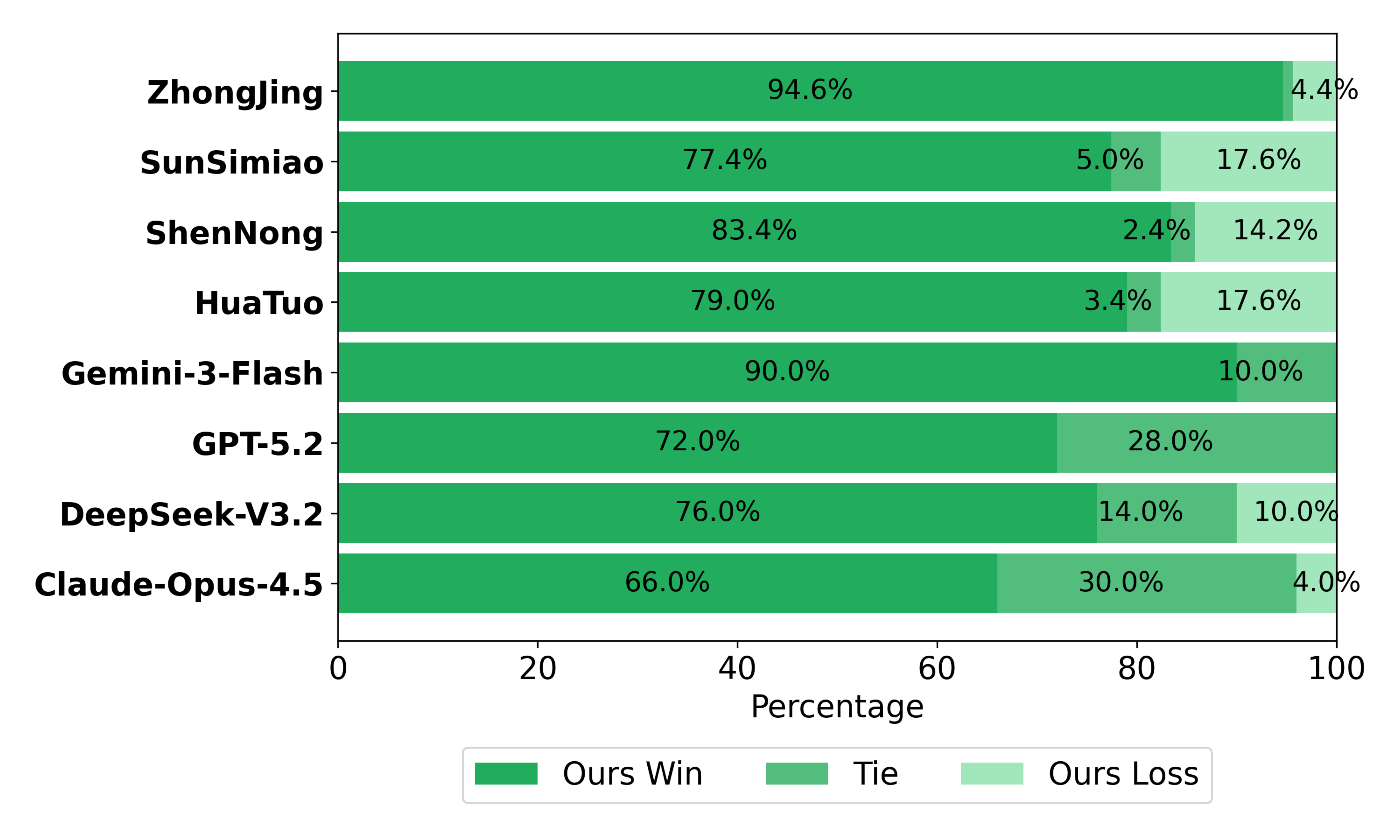}
  }\hfill
  \subfloat[Evaluation of Overall Effectiveness\label{fig:overall_effectiveness}]{
    \includegraphics[width=0.48\textwidth]{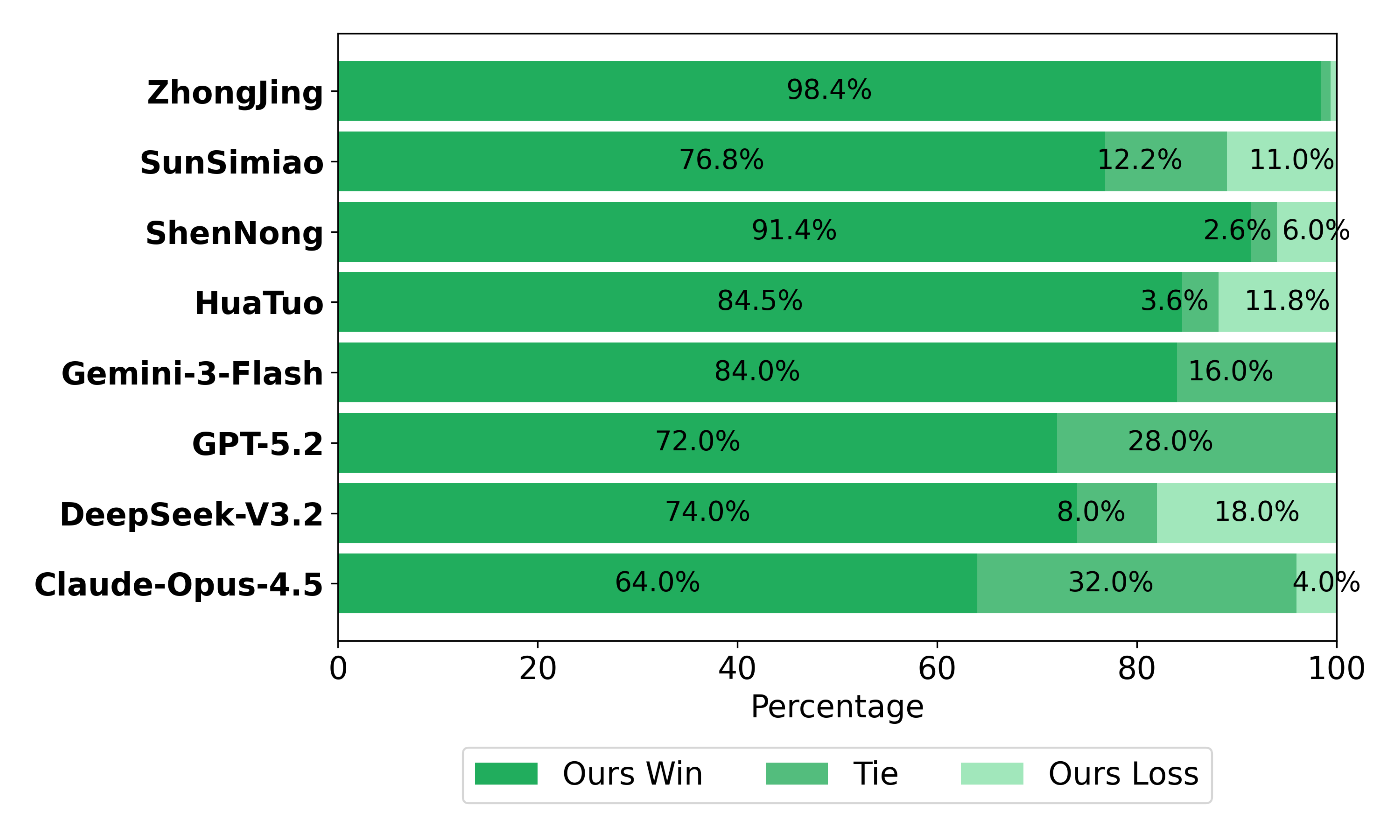}
  }
  \caption{ Comparison of multi-round consultation capabilities among different models.}
  \label{fig:evaluation_comparison}
\end{figure*}

To comprehensively assess the capabilities of the multi-round medical consultation among different models, 50 TCM cases were randomly selected from the database. These cases covered a range of clinical conditions, providing a comprehensive reflection of the diversity and complexity of TCM consultations. Based on these cases, an LLM acted as the patient, interacting with JF, the baseline models and the SOTA models to simulate authentic, multi-round clinical consultations and collect medical information for evaluation.

Five experienced TCM experts were invited to assess the performance of each model on four key dimensions: \textbf{Proactiveness}, \textbf{Accuracy}, \textbf{Practicality}, and \textbf{Overall Effectiveness}. Specifically: (1) \textbf{Proactiveness} measures whether the model can actively guide the consultation, adaptively adjusting the direction of questioning according to the patient's symptoms, and avoiding redundant or ineffective questions. (2) \textbf{Accuracy} evaluates the model’s ability to generate precise symptom-related questions that help patients express themselves clearly, reducing misunderstandings and biases. (3) \textbf{Practicality} examines whether the model can collect all the essential medical information (symptoms, medical history, lifestyle, etc.) needed for a clinical diagnosis and provide useful suggestions. (4) \textbf{Overall Effectiveness} focuses on how well the model integrates the information it has collected about patients, avoids irrelevant content, and supports efficient and effective diagnostic reasoning.  

Pairwise comparisons were conducted between the proposed JF system and each baseline of the TCM domain-specific model and the general SOTA model. Based on performance, each comparison was classified as \textit{win}, \textit{loss}, or \textit{tie}. The evaluation of each dimension focused on the performance of the model in critical consultation skills, comprehensiveness, and pertinence. As demonstrated in Figure ~\ref{fig:evaluation_comparison}, the developed Multi-Agent Collaborative Consultation Mechanism (MACCM) has been shown to significantly enhance the overall performance of the model in clinical medical consultation. A comparison of the JF with various baseline models reveals its consistent superiority in the four dimensions of proactivity, accuracy, practicality, and overall effectiveness. The results of the study indicate that the MACCM exhibits superior interaction and information processing capabilities during multi-round medical consultations. It is evident that, in terms of performance, MACCM provides a robust foundation for models to perform precise clinical diagnoses. Moreover, it is apparent that MACCM has considerable potential for real-world applications.

\subsubsection{Robustness of Multi-Round Medical Consultation}
\label{sec:robustness}

To further verify the robustness and reliability of the advantages of MACCM in expert-rated multi-round medical consultation identified by the JF system, a 95\% confidence interval analysis was conducted in four evaluation dimensions and eight comparison groups. This quantitative evaluation serves to complement the qualitative results illustrated in Fig.~\ref{fig:evaluation_comparison}. As demonstrated in Tab.~\ref{tab:ci_analysis}, JF consistently demonstrated robust and statistically stable performance across all criteria. The mean win rate was found to exceed 77\%, with no lower bound below 73\%, and the mean differences from the baselines were greater than 53\%. The 95\% confidence intervals for all variables did not include the value of zero, which confirms that the superiority of the MACCM in JF is not merely incidental, but rather robust. 

In particular, JF achieved an accuracy rate in excess of 91\% in the domains of \emph{Proactiveness} and \emph{Overall Effectiveness}. This demonstrated the efficacy of the proposed MACCM in improving both the initiative and the comprehensiveness of medical consultations. In the context of the subjective dimension of \emph{Practicality}, JF demonstrated an average success rate exceeding 85\%, thereby signifying a high degree of consistency in the model in various expert assessments. The findings indicate that the improvements of the model made by MACCM are statistically and practically significant, corroborating the reliability and universality of JF in diverse clinical scenarios.

\begin{table}[htbp]
\centering
\renewcommand{\arraystretch}{1.15}
\setlength{\tabcolsep}{3pt}
\begin{tabularx}{\linewidth}{
>{\centering\arraybackslash}X
>{\centering\arraybackslash}p{1.8cm}
>{\centering\arraybackslash}p{1.9cm}
>{\centering\arraybackslash}p{2.0cm}
}
\toprule
\textbf{Dimension} & 
\textbf{Avg. Win (\%)} & 
\textbf{Mean Diff. (\%)} & 
\textbf{95\% CI Range} \\
\midrule
Proactiveness & 86.75 & 83.75 & 79.25–88.00 \\
Overall Effect & 83.75 & 78.50 & 80.25–87.00 \\
Accuracy & 88.00 & 86.25 & 84.75–91.01 \\
Practicality & 86.75 & 84.00 & 83.49–90.00 \\
\bottomrule
\end{tabularx}
\caption{The 95\% confidence interval for the performance of JF on the four evaluation dimensions}
\label{tab:ci_analysis}
\end{table}

\subsubsection{Semantic and Lexical Consistency Analysis}
\label{sec:semantic}
In addition to the utilization of expert scoring, the \emph{semantic fidelity} and \emph{lexical consistency} of multi-round medical consultation records were subjected to further evaluation in order to assess the extent to which each model is able to reconstruct reference medical consultation. Tab.~\ref{semantic_bleu} provides a summary of the embedding-based semantic similarity (mean, standard deviation, minimum, maximum) and BLEU-1 lexical overlap scores, including the ablation variant of JF that does not incorporate the TCM General Agent.


\begin{table}[htbp]
\centering
\renewcommand{\arraystretch}{1.3}
\setlength{\tabcolsep}{3pt}
\scalebox{0.88}{
\begin{tabularx}{\linewidth}{cccccc}
\toprule
\textbf{Model} 
& \textbf{Mean Sim.} 
& \textbf{Std.} 
& \textbf{Min.} 
& \textbf{Max.} 
& \textbf{BLEU-1} \\
\midrule
Claude-Opus-4.5
  & 0.7711 & 0.0844 & 0.6119 & 0.9253 & 0.3156 \\
DeepSeek-V3.2
  & 0.7147 & 0.1206 & 0.4661 & 0.9087 & 0.1305 \\
Gemini-3-Flash
  & 0.8313 & 0.0757 & 0.5991 & 0.9392 & 0.2097 \\
GPT-5.2
  & 0.8275 & 0.0580 & 0.6841 & 0.9380 & 0.2720 \\
HuaTuoGPT2
  & 0.8213 & 0.0990 & 0.4795 & 0.9629 & 0.2125 \\
ShenNong
  & 0.8222 & 0.1012 & 0.4729 & 0.9453 & 0.1783 \\
SunSimiao
  & 0.7824 & 0.0869 & 0.4917 & 0.9043 & 0.1527 \\
ZhongJing
  & 0.7743 & 0.0927 & 0.5073 & 0.9141 & 0.1065 \\
\textbf{JF (Ours)}
  & \textbf{0.8409} & \textbf{0.0743} & \textbf{0.6040} & \textbf{0.9614} & \textbf{0.3305} \\
\textbf{JF w/o}
  & 0.8323 & 0.0818 & 0.5615 & 0.9487 & 0.1882 \\
\bottomrule
\end{tabularx}
}
\caption{Comparison of semantic similarity and BLEU-1 lexical overlap among different models, including the JingFang system without the TCM General Agent (denoted as JF w/o).}
\label{semantic_bleu}
\end{table}

The study demonstrated that JF exhibited the highest mean semantic similarity (\textbf{0.8409}) and the lowest variance (\textbf{0.0743}). This finding suggests that it exhibits superior alignment stability across the samples. Furthermore, it attains the maximum BLEU-1 score (\textbf{0.3305}), which signifies superior preservation of critical clinical terminology and phrasing. In contrast, the ablation variant of JF lacking the TCM General Agent demonstrates comparable semantic similarity (0.8323), but exhibits a marked decrease in BLEU-1 (0.1882), indicating reduced lexical precision and coverage. 

The findings demonstrate that the integration of the TCM General Agent into the system results in a substantial enhancement of its semantic coherence and linguistic accuracy. In conjunction with the confidence interval analysis, it is emphasized that JF's Multi-Agent Collaboration Consultation Mechanism (MACCM) not only enhances clinical reasoning and consultation robustness but also optimizes fidelity at both the semantic and lexical levels. This is of paramount importance for ensuring interpretable and trustworthy AI-assisted TCM consultation.

\subsection{Ablation Experiments}
\label{ablation experiment}
In order to assess the effectiveness of key components, i.e., Multi-Agent Collaboration Consultation Mechanism (MACCM), the Syndrome Agent and Dual-Stage Retrieval Scheme (DSRS), within the JF framework, a series of ablation studies were conducted, mainly focusing on the capabilities of multi-round medical consultation, syndrome differentiation and the corresponding treatment. By systematically comparing different modules within the framework and assessing their performance in real TCM cases, this experiment comprehensively validated the applicability and efficacy of the mechanism designed, the method developed, and the dataset constructed in the work.

\subsubsection{Evaluation of TCM General Agent in MACCM} 
The experiment was meticulously selected \textbf{100 exemplary TCM clinical cases} from a comprehensive set of medical data, with the objective of evaluating the contribution of the TCM General Agent in MACCM and the impact of the combination with the TCM Specialist Agents on multi-round medical consultations. To simulate multi-round interactions, an LLM-based Patient Agent is employed, which dynamically generates corresponding responses from a patient's perspective during the consultation, effectively mimicking real-world consultation scenarios. To assess performance, a comparative analysis was conducted, in which the consultation results were evaluated against the original medical case information. The evaluation is centered on two key aspects: \textbf{comprehensiveness}, defined as the extent to which multi-dimensional health information is covered, and \textbf{pertinence}, which is the ability to adapt to case-specific features. To ensure objectivity, GPT-4o is utilized as an evaluation instrument to automatically compare consultation outputs from JF and its ablation variant and score them based on the established criteria.

\begin{table}[htbp]
\centering
\setlength{\tabcolsep}{16pt}
\begin{tabular}{lll}
\toprule
Metric & JF & JF w/o \\
\midrule
\makecell[l]{Number of Selections out \\ of a Total of 100 Times} & 89 & 11 
\\
\makecell[l]{Average Consultation \\ Rounds per Patient} & 9.09 & 4.94 \\
\bottomrule
\end{tabular}
\caption{The performance of JingFang with TCM General Agent was compared with that of JingFang without the TCM General Agent (noted as JF w/o) in a total of 100 TCM clinical cases.}
\label{Ablation Results}
\end{table}

As shown in Tab.~\ref{Ablation Results}, the JF system with the TCM General Agent exhibits a substantial improvement in performance in multi-round medical consultations compared to its counterpart. The evaluation revealed a clear preference for the model with the TCM General Agent, which was selected \textbf{89 times out of a total of 100 } due to its comprehensive and pertinent nature. In contrast, the JF system without the TCM General Agent was selected only \textbf{ 11 times out of a total of 100 }, underscoring the crucial role of the TCM General Agent in optimizing consultation quality. Furthermore, the model with the TCM General Agent achieves an average of 9.09 consultation rounds per patient, which is nearly twice the 4.94 rounds of the one without it. This finding indicates that the incorporation of the TCM General Agent within the model enhances the comprehensiveness and scope of multi-round consultations, thereby markedly reinforcing the efficacy and performance of the JF in medical consultations.

\subsubsection{Applicability Assessment of the TCM Syndrome Differentiation Method} 
\label{Applicability Assessment of the TCM Syndrome Differentiation Dataset}

The objective of this ablation experiment is twofold: first, to demonstrate the effectiveness and applicability of the proposed method and the constructed data for TCM syndrome differentiation; and second, to explore the influence of different foundation models (or the model with a different size of parameters) on syndrome differentiation. Several representative open-source LLMs were selected for the purposes of evaluation, which include the Qwen-2.5 series (Qwen-2.5-3B-Instruct, Qwen-2.5-7B-Instruct, Qwen-2.5-14B-Instruct)~\cite{4}, the Qwen-3 series (Qwen-3-4B, Qwen-3-8B, Qwen-3-14B)~\cite{34}, the DeepSeek series (DeepSeek-7B-Chat)~\cite{deepseek-llm}, and the Llama series (Llama3.1-8B-Instruct)~\cite{2}, and the GLM series (GLM-4-9B Chat)~\cite{32}, and the Gemma series (Gemma-3-12b-it)~\cite{33}.

\begin{figure}[htbp] 
\includegraphics[width=1\columnwidth]{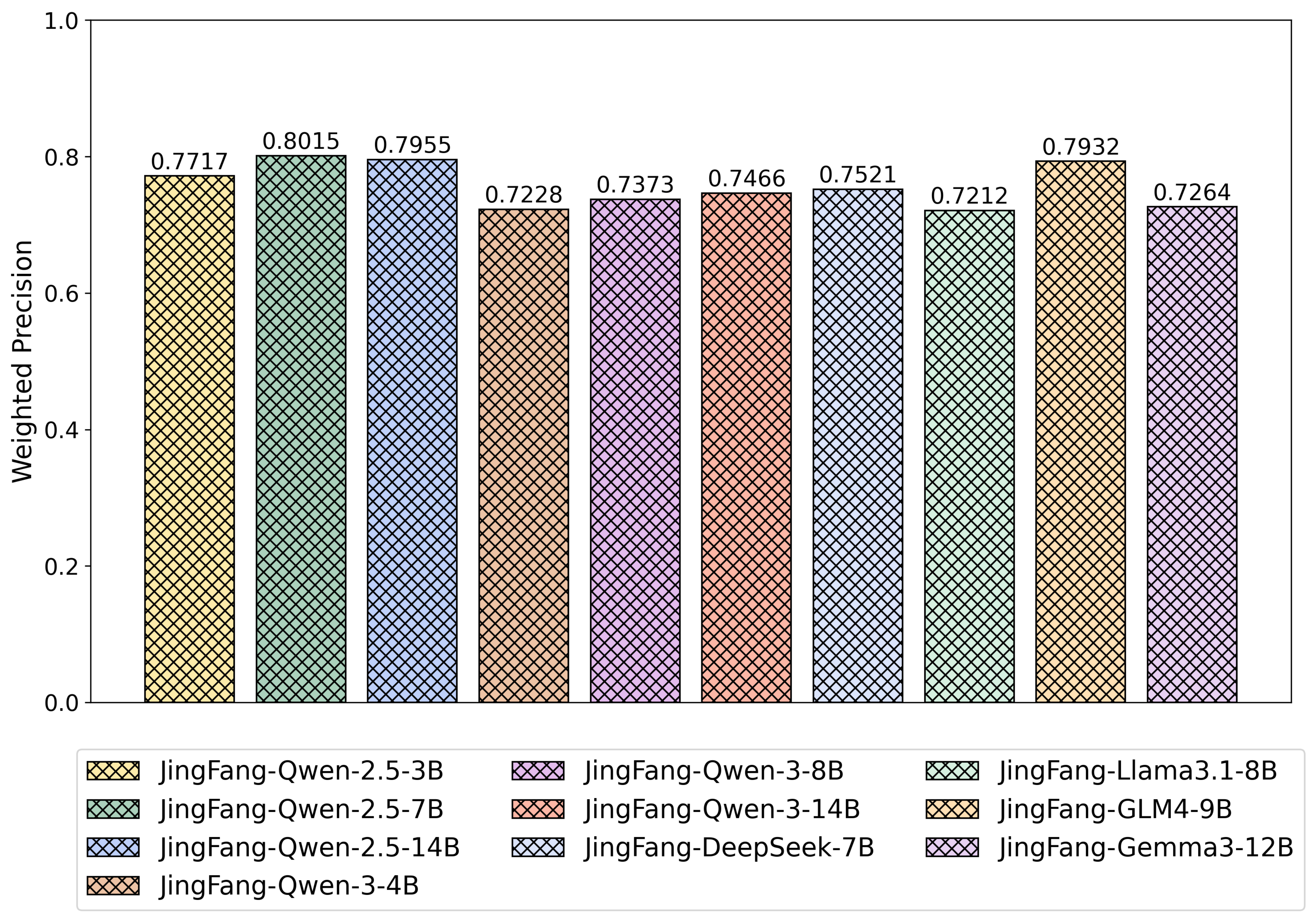}
\caption{An ablation experiment on syndrome differentiation precision of different foundation models that are integrated with the developed multi-agent system, which demonstrates the effectiveness and applicability of the developed method and dataset in this work.}
\label{TCM Syndrome Differentiation Dataset}
\end{figure}

As shown in Fig.~\ref{TCM Syndrome Differentiation Dataset}, all models in the experiment exhibited advanced performance in TCM syndrome differentiation compared to their original versions after being trained using our method and data. The findings indicate that the selection of the foundation model is not the primary factor that contributes to the enhancement of the capability of syndrome differentiation; rather, the methodology and the data play a more significant role. Moreover, the findings illustrate the prospective relevance and worth of the proposed TCM syndrome differentiation approach and dataset for subsequent research, particularly in light of its superior performance compared to that of GPT-5.2 (0.6323 in precision). 

\subsubsection{Ablation Study on the Dual-Stage Retrieval Scheme}
To verify the effectiveness of the Dual-Stage Retrieval Scheme (DSRS) developed for the Treatment Agent, a comparative experiment was designed and conducted to compare DSRS with the direct single-stage symptom-matching retrieval approach (denoted as Single-Stage approach) . In the experiment, a total of \textbf{100 clinical cases} were randomly selected. Under a unified recall threshold, both the DSRS and the Single-Stage approach were applied to retrieve each case. The similarity between the retrieved results obtained from each method and the original case was then measured and evaluated.

\begin{figure}[htbp]
\vspace{0.2em}
\centering
\includegraphics[width=\columnwidth]{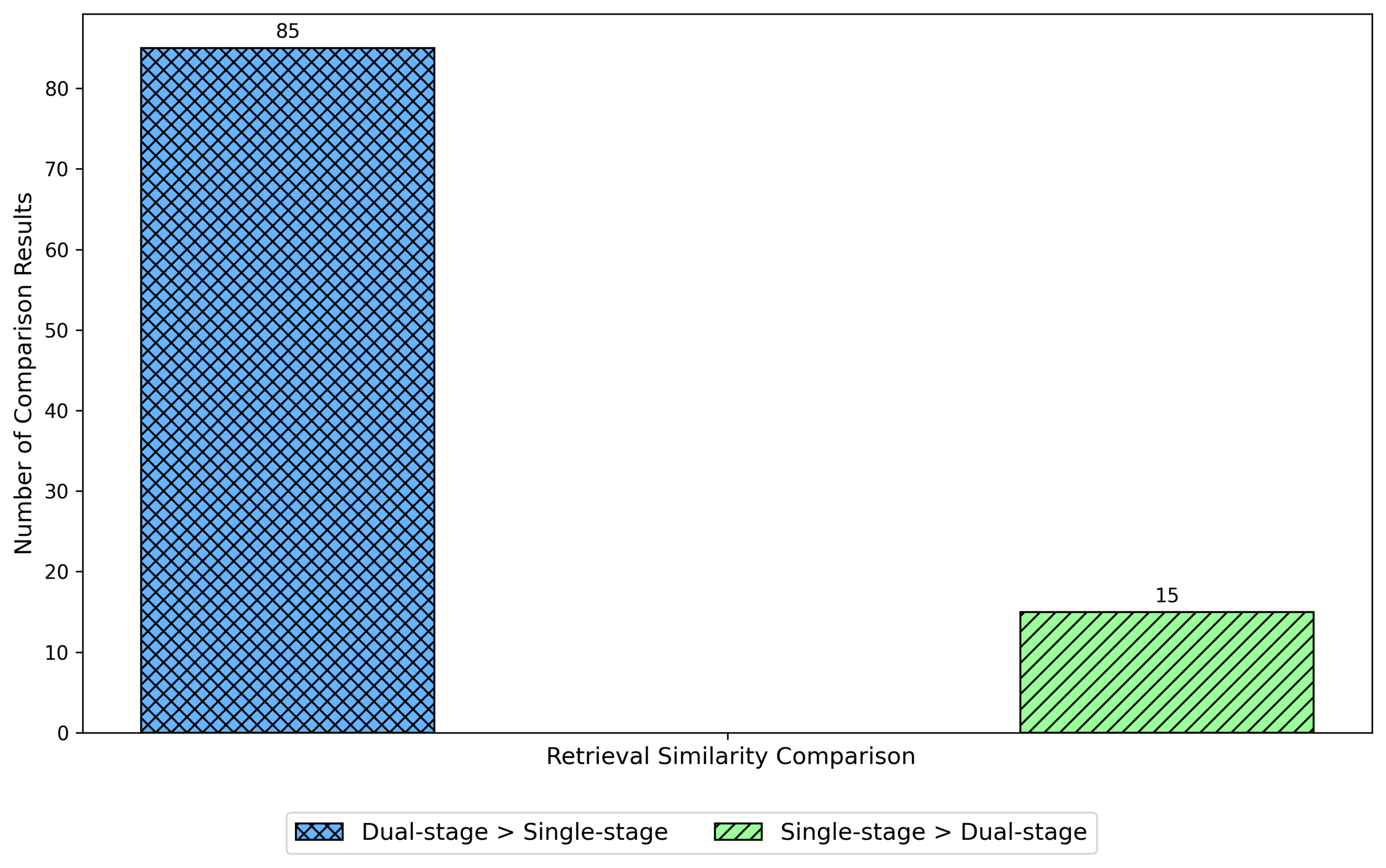}
\caption{Comparison result of the retrieval similarity between the DSRS and the Single-Stage approach in a total of 100 TCM clinical cases.}
\label{dsrs_ablation}
\vspace{-0.2em}
\end{figure}

As shown in Fig.~\ref{dsrs_ablation}, the DSRS demonstrates a marked improvement in performance compared to the single-stage approach, which is closely aligned with the original cases. In the test involving 100 clinical cases, the DSRS achieved significantly higher similarity scores than the Single-Stage approach in 85 cases, with lower scores observed in only 15 cases. Subsequent analysis indicates that DSRS effectively reduces semantic interference from non-target syndromes with similar symptoms, enhancing both the specificity and reliability of retrieval. This mechanism facilitates a more precise comprehension of fundamental TCM principles, including the concept of ``different treatments for the same disease'' and ``the same treatment for different diseases'' . Consequently, it enables for more precise and personalized clinical treatment recommendations.

\subsection{Case Studies}
\begin{figure}[htbp]
  \centering
  \includegraphics[width=\columnwidth]{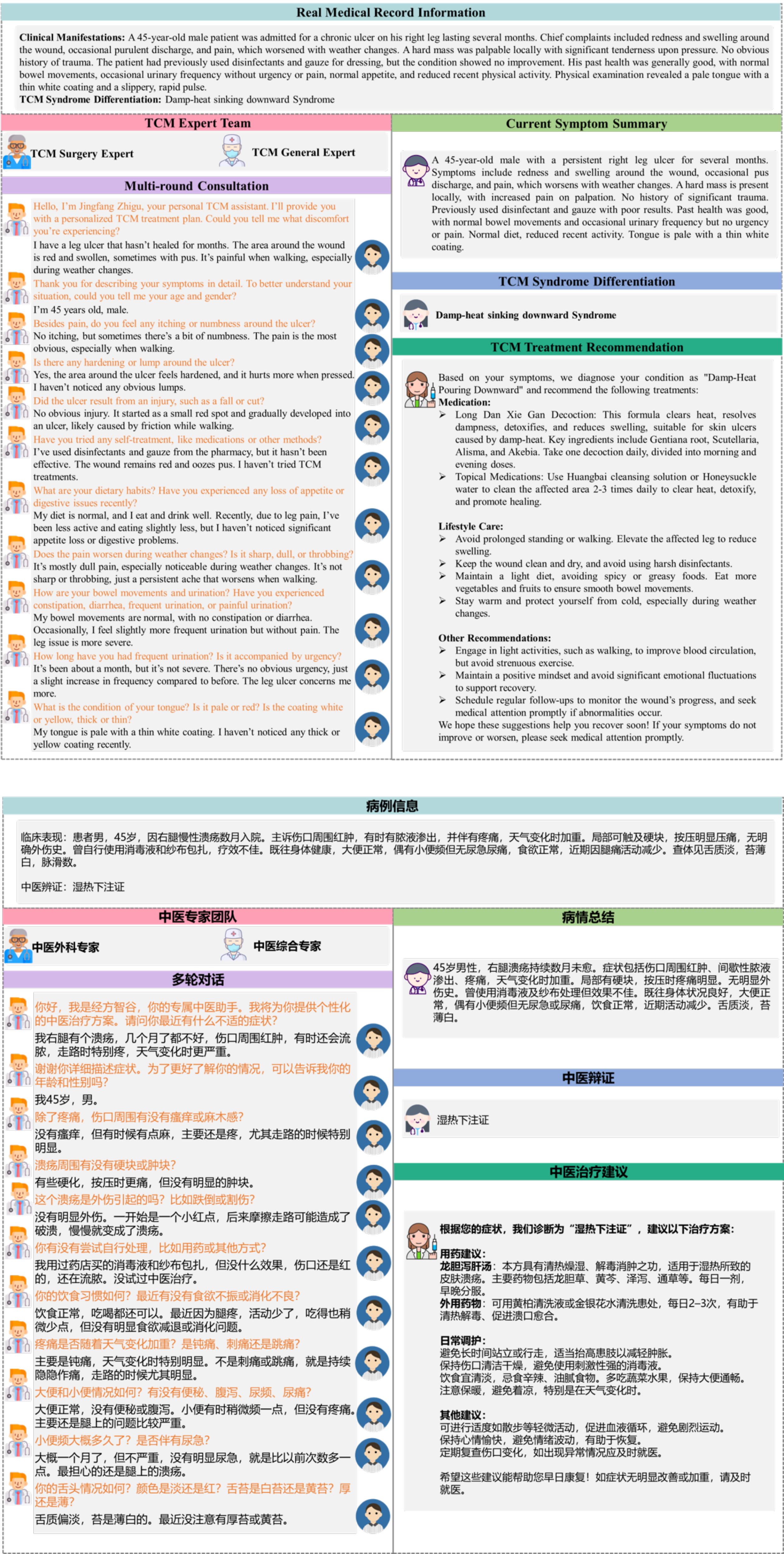}
  \caption{Case study of Damp-Heat Sinking Downward Syndrome.}
  \label{fig:damp_heat_case}
\end{figure}

\begin{figure}[htbp]
  \centering
  \includegraphics[width=\columnwidth]{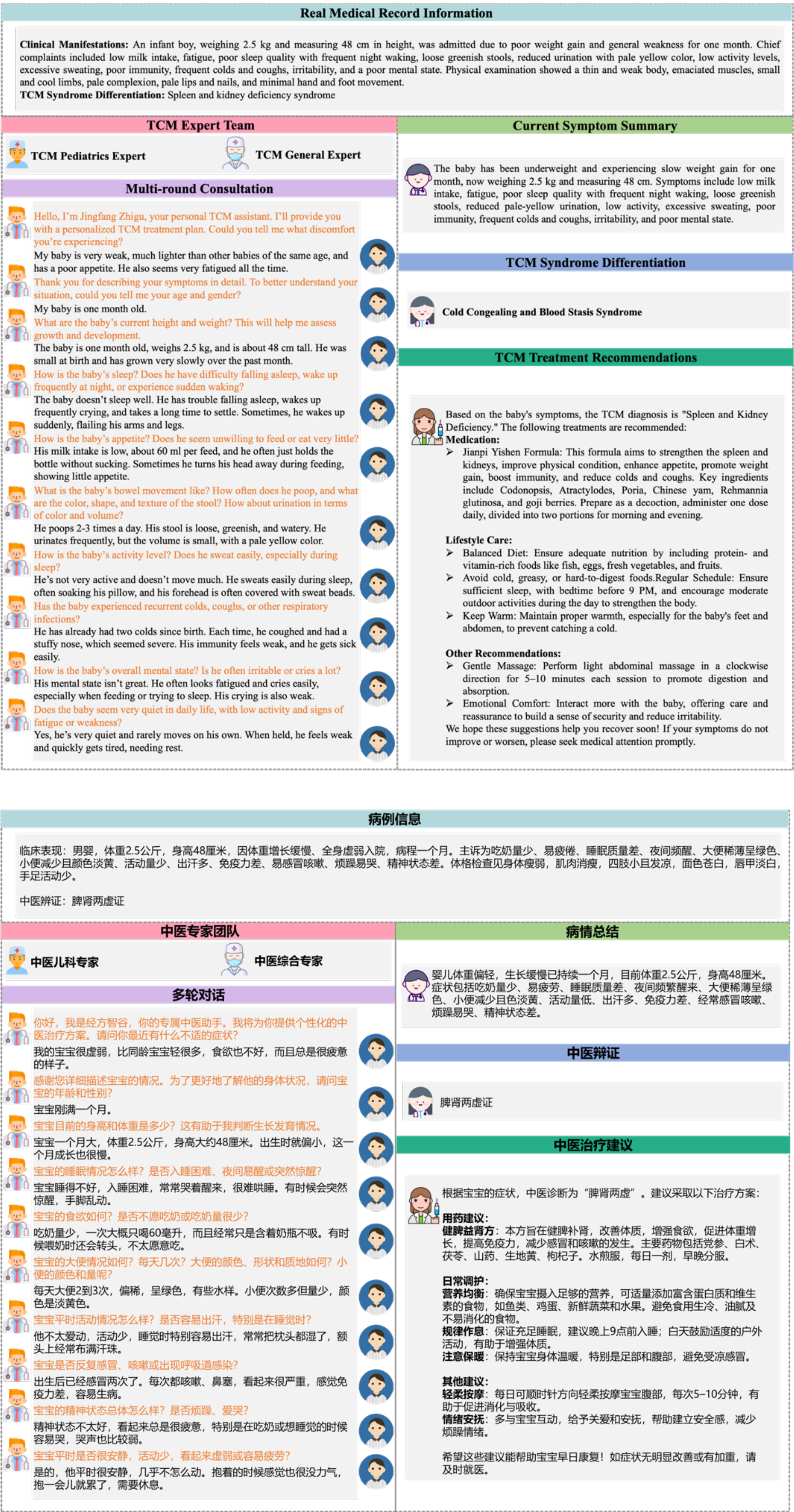}
  \caption{Case study of Pediatric Spleen--Kidney Yang Deficiency Syndrome.}
  \label{fig:spleen_kidney_case}
\end{figure}

Simulation experiments were conducted based on typical cases extracted from classical TCM literature, exhibiting the applicability and interpretability of the \textbf{JingFang (JF)} model with the proposed method in authentic clinical scenarios. The model demonstrated stable, reliable, and TCM-consistent diagnostic and reasoning capabilities across different diseases through multi-round interactive consultation, syndrome differentiation, and treatment recommendation. These capabilities were evident in the model's performance on diseases such as \emph{Damp-Heat Sinking Downward Syndrome} and \emph{Spleen--Kidney Yang Deficiency Syndrome}. The results of the experiment are illustrated in Fig.~\ref{fig:damp_heat_case} and Fig.~\ref{fig:spleen_kidney_case}.

The results of the case study indicate three primary benefits of the designed framework for JingFang. First, the \textbf{MACCM} facilitates proactive and comprehensive inquiry, ensuring efficient information collection during consultation. Second, the \textbf{Syndrome Agent} employs reasoning chains that are transparent and interpretable, according to the diagnostic logic of traditional Chinese medicine (TCM). Third, the \textbf{DSRS} enables the Treatment Agent to formulate customized and syndrome-consistent prescriptions, thereby establishing a closed-loop workflow that encompasses consultation, syndrome differentiation, and treatment recommendation, as illustrated in Fig.~\ref{fig:damp_heat_case} and Fig.~\ref{fig:spleen_kidney_case}. This integrated process provides a practical and interpretable path toward AI-assisted TCM diagnosis and treatment.

\section{Conclusion and Discussion}
In this article, we present the development of JingFang (JF), an novel LLM-based multi-agent system for TCM that exhibits expertise-level capabilities, particularly in providing accurate medical consultation, diagnosis, and syndrome differentiation. To this end, we have meticulously developed the multi-agent system that allows various TCM Specialist Agents collaborate to emulate and align with authentic TCM diagnosis and treatment scenarios. The proposed Multi-Agent Collaboration Consultation Mechanism (MACCM) endows the model with comprehensive and targeted consultation capabilities and facilitates explicit reasoning and decision-making during the consultation process. This, in turn, provides a solid foundation for accurate diagnosis and syndrome differentiation. The experimental results verify that JF outperforms current TCM-specific models and SOTA LLMs in terms of multi-round consultation based on four professional evaluation dimensions. Furthermore, a TCM Syndrome Agent is calibrated with the preprocessed and structured TCM dataset and combines with the designed Dual-Stage Recovery Scheme (DSRS) to significantly improve the interpretability and accuracy of base models in the differentiation and treatment of the syndrome. The experimental results also demonstrate that the precision of syndrome differentiation has been improved by at least 124\% and 21.1\% by JF, respectively, compared to the existing TCM models and the SOTA LLMs. 

JingFang overcomes the key limitations of current TCM models, thereby improving the effective applications of LLM in the TCM domain and facilitating the implementation of AI-assisted TCM. In the future, due to the complexity of real clinical scenarios in TCM, further exploration of Multi-Agent Collaboration Mechanisms and development of advanced multimodal diagnostic technologies to establish more precise and professional TCM models will be of great practical significance.

\bibliography{reference}
\bibliographystyle{elsarticle-harv}

\end{document}